\documentclass[a4paper,11pt]{article}
\usepackage{fourier}
\usepackage{color}
 \usepackage{graphicx}
\usepackage{url}
\usepackage{amsmath}
\usepackage{wrapfig}
\usepackage{xspace}

\usepackage{array}
\usepackage{subcaption}
\usepackage{multicol}
\usepackage{svg}
\usepackage{breakcites}

\usepackage[T1]{fontenc}
\usepackage{times}
\usepackage{hyperref}
\usepackage[nameinlink]{cleveref}
\usepackage{orcidlink}

\hypersetup{
    colorlinks = true,
    citecolor = blue
}

\newcommand\blfootnote[1]{%
  \begingroup
  \renewcommand\thefootnote{}\footnote{#1}%
  \addtocounter{footnote}{-1}%
  \endgroup
}

\title{Fast and Efficient Scene Categorization \\ for Autonomous Driving using VAEs}
\author{Saravanabalagi Ramachandran \orcidlink{0000-0001-6543-5345}$^1$, Jonathan Horgan$^2$, Ganesh Sistu$^2$, and John McDonald \orcidlink{0000-0001-9225-673X}$^1$}

\date{%
    \textit{$^1$Department of Computer Science, Maynooth University, Ireland}\\
    \textit{$^2$Valeo Vision Systems, Ireland}
}

\begin{document}
\maketitle
\thispagestyle{empty}

\begin{abstract}
Scene categorization is a useful precursor task that provides prior knowledge for many advanced computer vision tasks with a broad range of applications in content-based image indexing and retrieval systems. Despite the success of data driven approaches in the field of computer vision such as object detection, semantic segmentation, etc., their application in learning high-level features for scene recognition has not achieved the same level of success. We propose to generate a fast and efficient intermediate interpretable generalized global descriptor that captures coarse features from the image and use a classification head to map the descriptors to 3 scene categories: Rural, Urban and Suburban. We train a Variational Autoencoder in an unsupervised manner and map images to a constrained multi-dimensional latent space and use the latent vectors as compact embeddings that serve as global descriptors for images. The experimental results evidence that the VAE latent vectors capture coarse information from the image, supporting their usage as global descriptors. The proposed global descriptor is very compact with an embedding length of 128, significantly faster to compute, and is robust to seasonal and illuminational changes, while capturing sufficient scene information required for scene categorization. 

\blfootnote{$^{1}${\scriptsize\{saravanabalagi.ramachandran, john.mcdonald\}@mu.ie}. $^{2}${\scriptsize\{jonathan.horgan, ganesh.sistu\}@valeo.com}. This research was supported by \href{https://www.sfi.ie}{Science Foundation Ireland} grant 13/RC/2094 to \href{https://www.lero.ie}{Lero - the Irish Software Research Centre} and grant 16/RI/3399.}

\end{abstract}


\section{Introduction}
\label{Introduction}

\begin{wrapfigure}{r}{0.5\textwidth}
    \centering
    \includegraphics[width=0.5\textwidth]{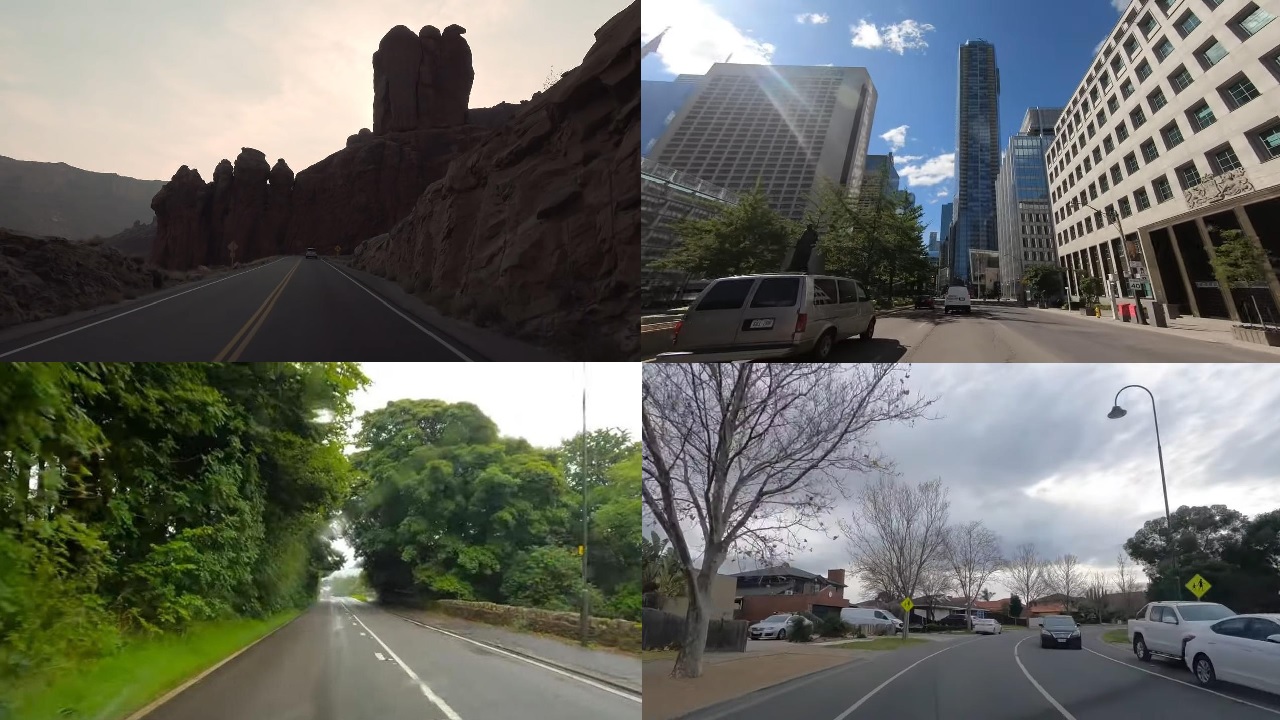}
    \caption{Images from the Scene Categorization Dataset. Top Left: Rural (Utah), Top Right: Urban (Toronto), Bottom Left: Rural (Stockport to Buxton), Bottom Right: Suburban (Melbourne)}.
    \label{fig:scene_categorization}
    \vskip -20pt
\end{wrapfigure}

Scene categorization is a precursor task with a broad range of applications in content-based image indexing and retrieval systems. Content Based Image Retrieval (CBIR) uses the visual content of a given query image to find the closest match in a large image database \cite{dl_holistic_feature_descriptors_9287063}. The retrieval accuracy of CBIR depends on both the feature representation and the similarity metric. The retrieval process can be accelerated by selectively searching based on certain scene categories e.g. given a query image with multiple high-rise buildings, searching the rural regions would not be beneficial and can be skipped. The knowledge about the scene category can also assist in context-aware object detection, action recognition, and scene understanding and provides prior knowledge for other advanced computer vision tasks \cite{indoorSceneRecognition_khan2016discriminative,sunDatabaseSceneRecognition_xiao2010sun}. 

In autonomous driving scenarios, location context provides an important prior for parameterising autonomous behaviour. Generally GPS data is used to determine if the vehicle has entered the city limits, where additional caution is required 
e.g. to set the pedestrian detection threshold to watch out for pedestrians in populated regions. However, such an approach requires apriori labelling of the environment and due to rapid development of regions around the cities and suburbs, it has become increasingly hard to distinguish such regions of interest only using GPS coordinates. 
A more scalable and lower cost approach would be to
automatically determine the scene type at the edge using locally sensed data.

We present a deep learning based unsupervised holistic approach that directly encodes coarse information in the multi-dimensional latent space without explicitly recognizing objects, their semantics or capturing fine details. Models equipped with intermediate representations train faster, achieve higher task performance, and generalize better to previously unseen environments \cite{intermediateRepresentations_zhou2019does}. To this end, rather than directly mapping the input image to the required scene categories as with classic data-driven classification solutions, we propose to generate an intermediate generalized global descriptor that captures coarse features from the image and use a separate classification head to map the descriptors to scene categories. More specifically, we use an unsupervised convolutional Variational Autoencoder (VAE) to map images to a multi-dimensional latent space. We propose to use the latent vectors directly as global descriptors, which are then mapped to 3 scene categories: Rural, Urban and Suburban, using a supervised classification head that takes in these descriptors as input.
\section{Background}
\label{Background}



The success of deep learning in the field of computer vision over the past decade has resulted in dramatic improvements in performance in areas such as object recognition, detection, segmentation, etc. However, the performance of scene recognition is still not sufficient to some extent because of complex configurations \cite{sceneRecognitionSurvey_XIE2020107205}. Early work on scene categorization includes \cite{modelingshapeofscene_oliva2001modeling} where the authors proposed a computational model of the recognition of real world scenes that bypasses the segmentation and the processing of individual objects or regions. 
Notable early global image descriptor approaches include aggregation of local keypoint descriptors through Bag of Words (BoW)  \cite{bow_bovw_csurka2004visual}, Fisher Vectors (FV) \cite{fv1_fisher_perronnin2010improving,fv2_fisher_sanchez:hal-00830491} and Vector of Locally Aggregated Descriptors (VLAD) \cite{vlad_jegou2010aggregating}. More recently, researchers have also used Histogram of Oriented Gradients (HOG) and its extensions such as
Pyramid HOG (PHOG) for mapping and localization \cite{htmap_7938750}. Although these approaches have shown strong performance in constrained settings, they lack the repeatability and robustness required to deal with the challenging variability that occurs in natural scenes caused due to different times of the day, weather, lighting and seasons \cite{nuimeprn10987}.

To overcome these issues recent research has focussed on the use of learned global descriptors. Probably the most notable here is NetVLAD which reformulated VLAD through the use of a deep learning architecture~\cite{arandjelovic2016netvlad} resulting in a CNN based feature extractor using weak supervision to learn a distance metric based on the triplet loss.

Variatonal Autoencoder (VAE), introduced by \cite{vaniallaVae_kingma2013auto}, maps images to a multi-dimensional standard normal latent space.
Although since the introduction of the CelebA dataset \cite{celeba_liu2015deep} multiple implementations of VAEs have shown success in generating human faces, VAEs often produce blurry and less saturated reconstructions and have been shown to lack the ability to generalize and generate high-resolution images for domains that exhibit multiple complex variations e.g. realistic natural landscape images. 
Besides their use as generative models, VAEs have also been used to infer one or more scalar variables from images in the context of Autonomous Driving such as for vehicle control \cite{vaeendtoend_amini2018variational}.

 A number of researchers have developed datasets to accelerate progress in general scene recognition. Examples include MIT Indoor67 \cite{mitIndoor67k_quattoni2009recognizing}, 
SUN \cite{sunDatabaseSceneRecognition_xiao2010sun}, 
and Places 365 \cite{places365_zhou2017places}. Whilst these datasets capture a very wide variety of scenes they lack suitability when developing scene categorisation techniques that are specific to autonomous driving. Given this, in our research we choose to use images from public driving datasets such as Oxford Robotcar \cite{oxfordrobotcar_maddern20171} in an unsupervised manner 
and curate our own evaluation dataset targeted at our domain of interest.

In this paper, we propose to use a trained Variational Autoencoder to map images to a multi-dimensional latent space and use the latent vectors as compact embeddings that serve directly as global descriptors for images. To the best of our knowledge, this is the first time VAE latent vectors are used as global image descriptors. In detail, we train a convolutional Variational Autoencoder in an unsupervised manner with images from Oxford Robotcar dataset \cite{oxfordrobotcar_maddern20171} that exhibit strong visual changes caused by seasons, weather, time of the day, etc. and use the latent vectors inferred using the encoder as global descriptors. We show that VAE encoder captures coarse features of the image and produces a mapping in multi-dimensional standard normal latent space. We then use a simple two-layer linear classification perceptron head to map the global descriptors to required scene categories: Rural, Urban and Suburban.


\section{Methodology}
\label{Methodology}
In our method, we train a Variational Autoencoder from scratch on images from publicly available Oxford Robotcar dataset \cite{oxfordrobotcar_maddern20171} in an unsupervised manner. We then obtain latent vectors during inference and use them as compact embeddings that serve as global descriptors.
The global descriptors are then used as input to a simple linear classifier that maps them to 3 scene categories: Rural, Urban and Suburban.  \autoref{fig:arch} provides a high-level view of our overall architecture.


\begin{center}
    \includegraphics[width=\textwidth]{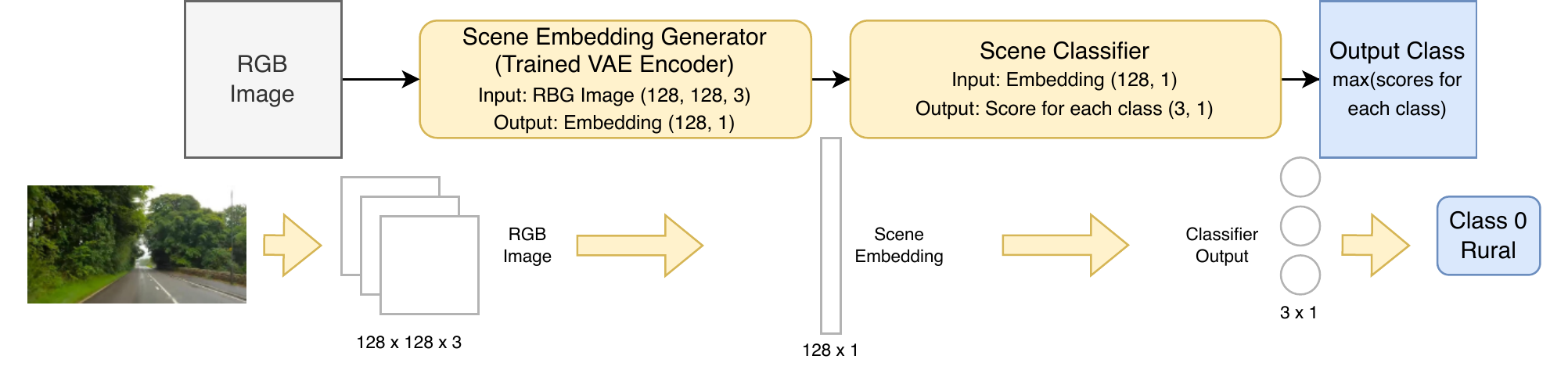}
    \captionof{figure}{Overall system architecture for scene categorization. Input and output for each module is shown in the bottom row including an example.}
    \label{fig:arch}
\end{center}

\subsection{Scene Embedding}
In order to evaluate the performance of different VAEs for scene embedding, we train multiple variants on 3 Oxford Robotcar traversals \textit{2014-12-09-13-21-02} (Winter Day), \textit{2014-12-10-18-10-50} (Winter Night), and \textit{2015-05-19-14-06-38} (Summer Day). These traversals exhibit changes due to seasons and time of the day. 
We sub-sample each traversal using the sequential adaptive sampling strategy reported in \cite{odoviz_9564712} with parameters $\tau_{d_{\text{acc}}} = 5 m$ and  $\tau_{\theta_{\text{acc}}} = 15^{\circ}$ to obtain 1787, 1879 and 1825 visually dissimilar images respectively. 
The following variants of VAE are trained:
\begin{itemize}
    \begin{multicols}{2}
    \itemsep -4pt
    \item BetaVAE                                  \cite{betavae_DBLP:conf/iclr/HigginsMPBGBML17}
    \item CategoricalVAE                           \cite{categoricalvae_jang2016categorical}
    \item DFCVAE                                   \cite{dfcvae_hou2017deep}
    \item DIPVAE                                   \cite{dipvae_kumar2017variational}
    \item InfoVAE                                  \cite{infovae_DBLP:journals/corr/ZhaoSE17b}
    \item LogCoshVAE                               \cite{logcoshvae_chen2019log}
    \item MIWAE                                    \cite{miwae_rainforth2018tighter}
    \item VAE (Vanilla)                            \cite{vaniallaVae_kingma2013auto}
    \end{multicols}
\end{itemize}

\begin{figure}
    \centering
    \captionsetup[subfigure]{format=hang,justification=raggedright}
    \begin{subfigure}{0.32\linewidth}
        \centering
        \includegraphics[width=\textwidth,trim={0 21cm 0 0},clip]{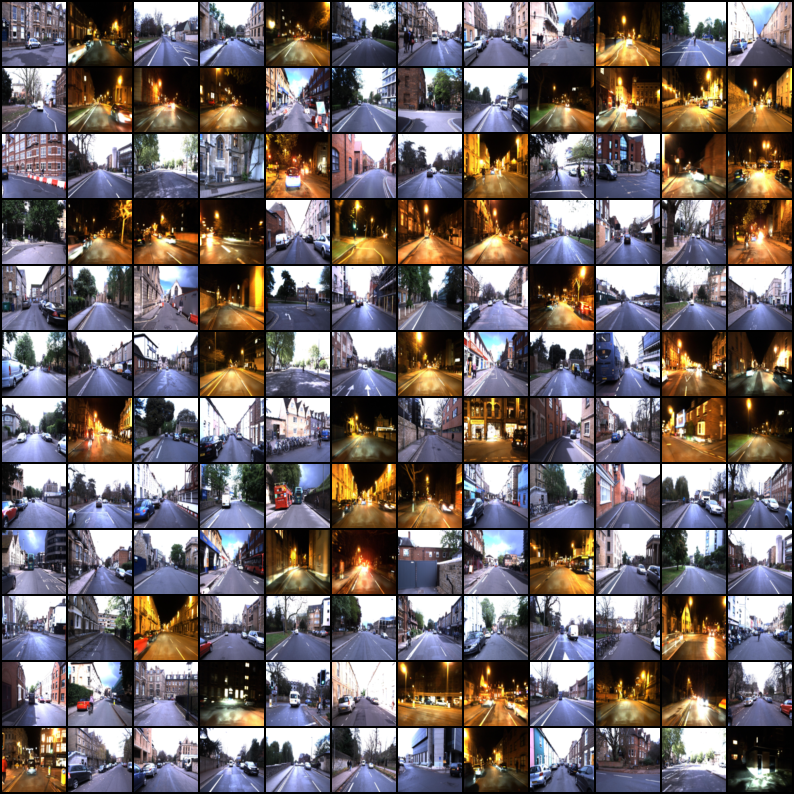}
        \caption{48 random input images from Oxford Robotcar dataset}
    \end{subfigure}
    \begin{subfigure}{0.32\linewidth}
        \centering
        \includegraphics[width=\textwidth,trim={0 21cm 0 0},clip]{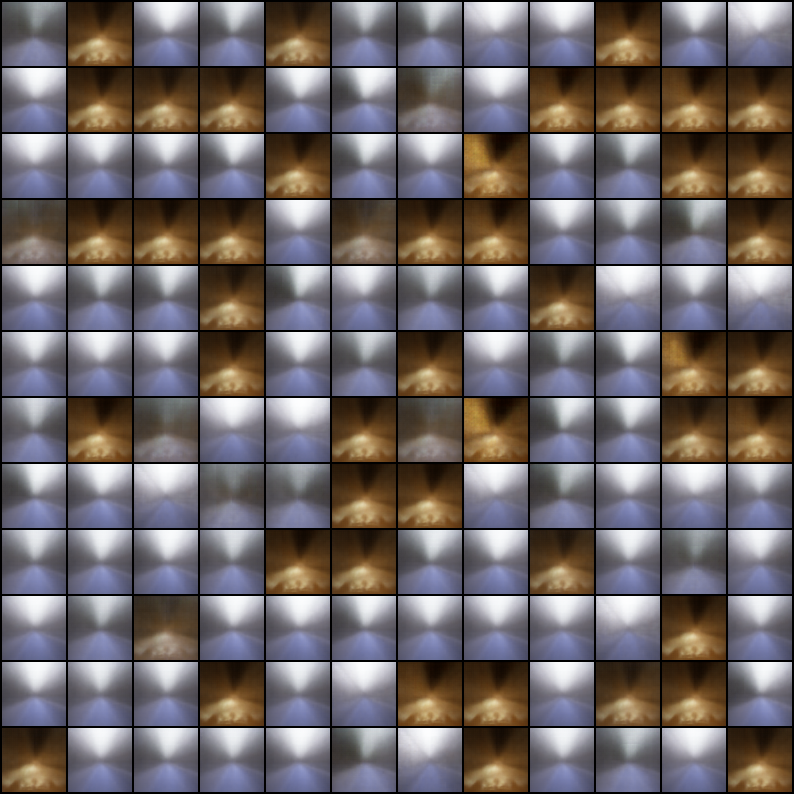}
        \caption{VAE (499 Epochs) \cite{vaniallaVae_kingma2013auto}}
    \end{subfigure}
    \begin{subfigure}{0.32\linewidth}
        \centering
        \includegraphics[width=\textwidth,trim={0 21cm 0 0},clip]{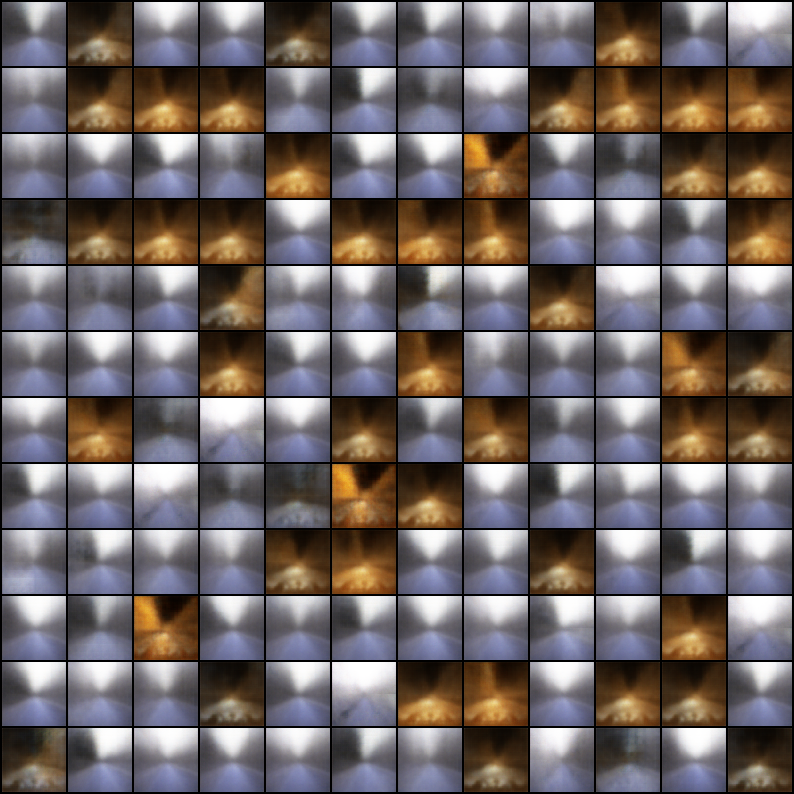}
        \caption{BetaVAE (103 Epochs) \cite{betavae_DBLP:conf/iclr/HigginsMPBGBML17}}
    \end{subfigure}
    
    \bigskip
    
    \begin{subfigure}{0.32\linewidth}
        \centering
        \includegraphics[width=\textwidth,trim={0 21cm 0 0},clip]{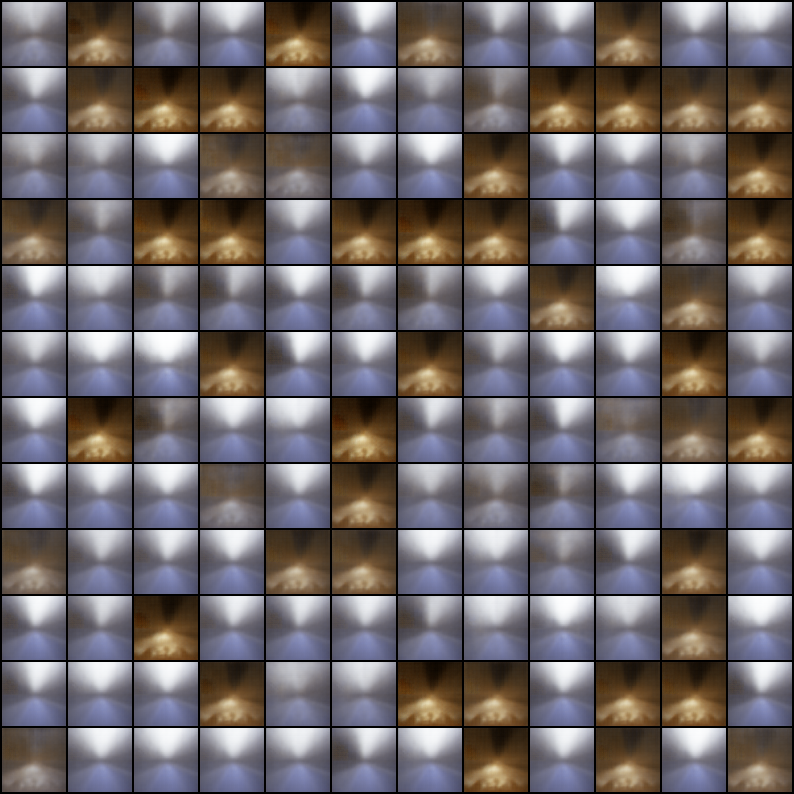}
        \caption{CategoricalVAE (114 Epochs) \cite{categoricalvae_jang2016categorical}}
    \end{subfigure}
    \begin{subfigure}{0.32\linewidth}
        \centering
        \includegraphics[width=\textwidth,trim={0 21cm 0 0},clip]{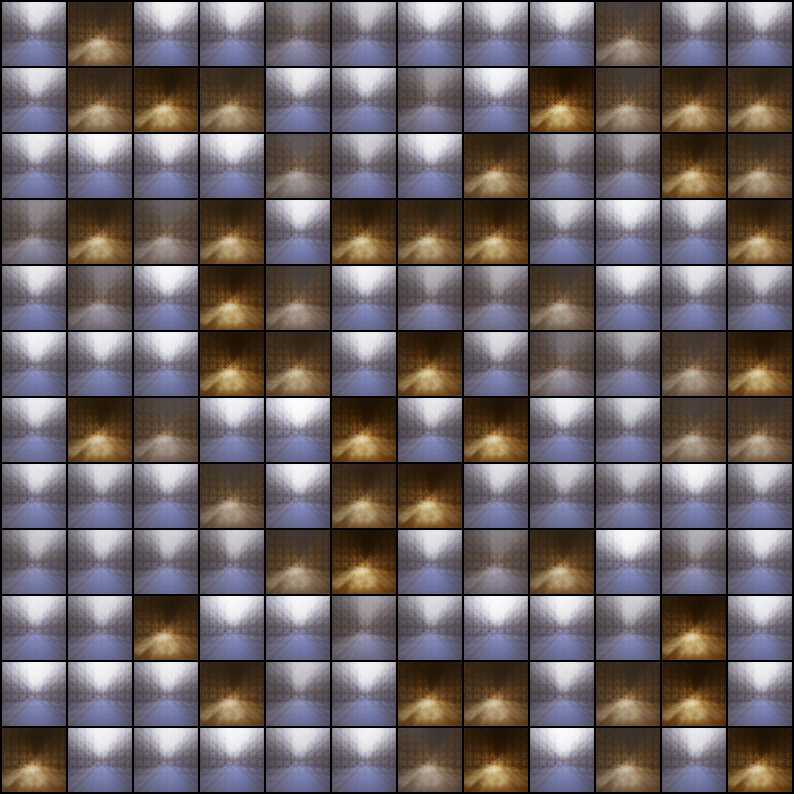}
        \caption{DFCVAE (107 Epochs) \cite{dfcvae_hou2017deep}}
    \end{subfigure}
    \begin{subfigure}{0.32\linewidth}
        \centering
        \includegraphics[width=\textwidth,trim={0 21cm 0 0},clip]{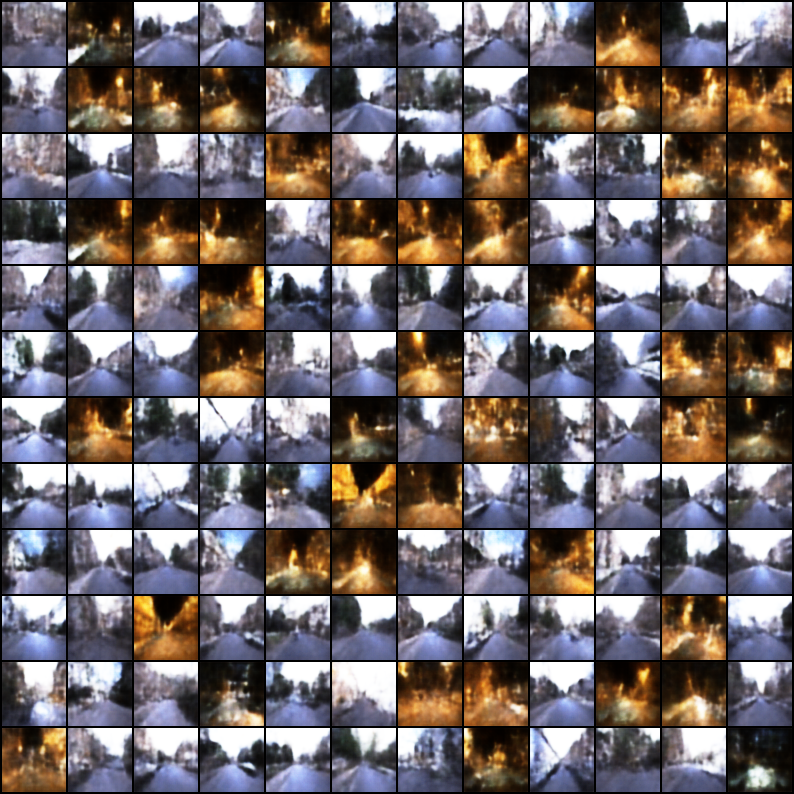}
        \caption{DIPVAE (178 Epochs) \cite{dipvae_kumar2017variational}}
    \end{subfigure}
    
    \bigskip
    
    \begin{subfigure}{0.32\linewidth}
        \centering
        \includegraphics[width=\textwidth,trim={0 21cm 0 0},clip]{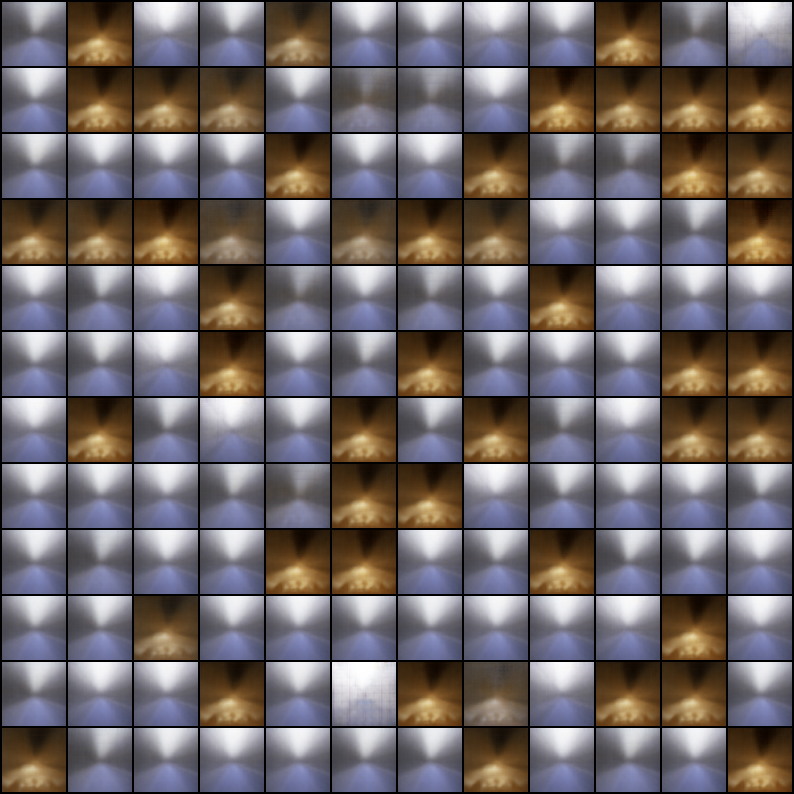}
        \caption{InfoVAE (119 Epochs) \cite{infovae_DBLP:journals/corr/ZhaoSE17b}}
    \end{subfigure}
    \begin{subfigure}{0.32\linewidth}
        \centering
        \includegraphics[width=\textwidth,trim={0 21cm 0 0},clip]{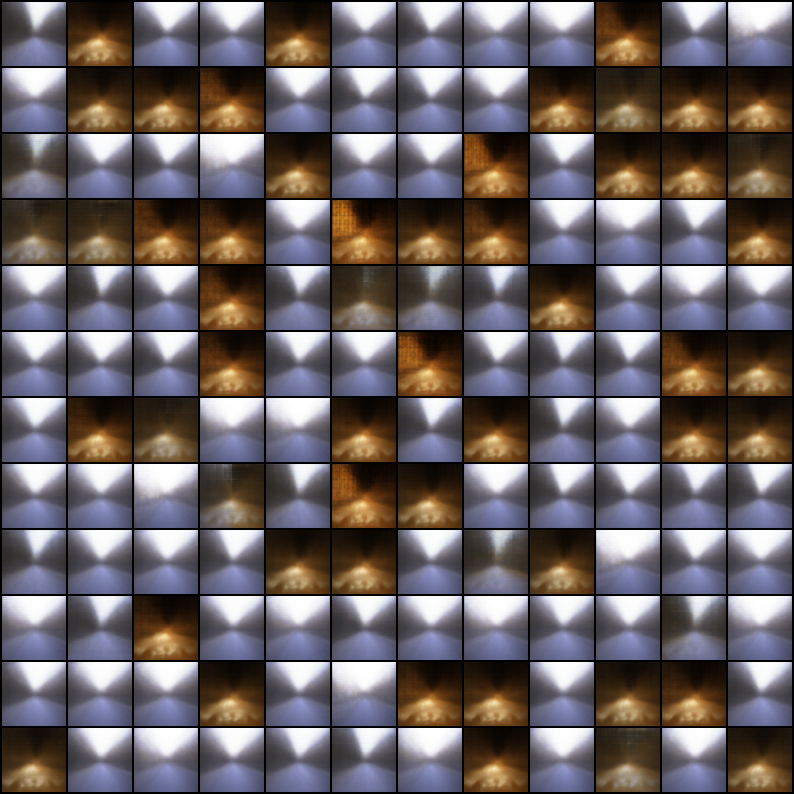}
        \caption{LogCoshVAE (104 Epochs) \cite{logcoshvae_chen2019log}}
    \end{subfigure}
    \begin{subfigure}{0.32\linewidth}
        \centering
        \includegraphics[width=\textwidth,trim={0 21cm 0 0},clip]{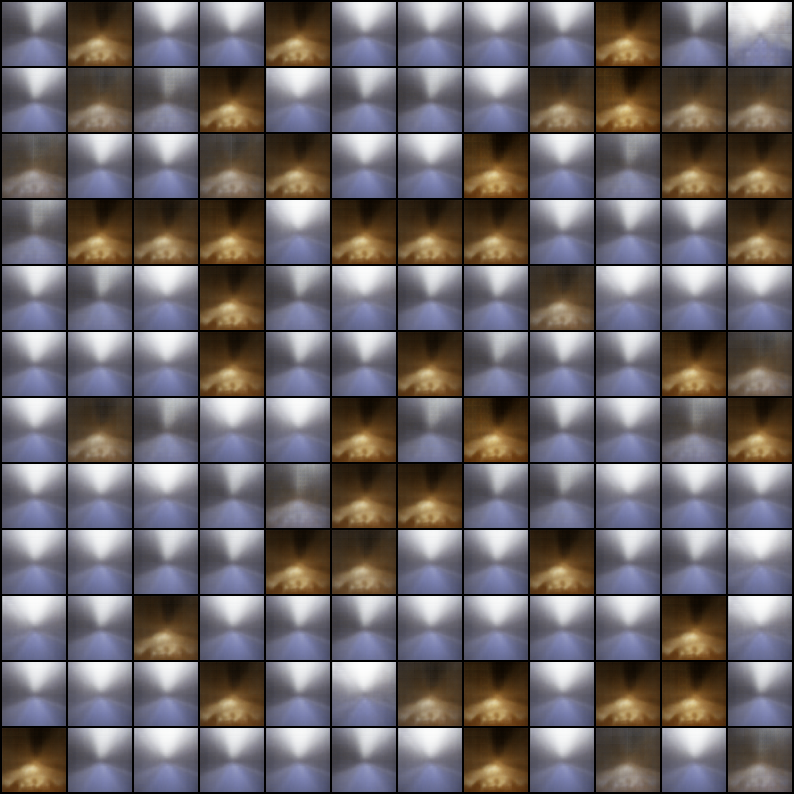}
        \caption{MIWAEVAE (103 Epochs) \cite{miwae_rainforth2018tighter}}
    \end{subfigure}
    \caption{Reconstructions of different variants of VAE}
    \label{fig:vae_recons}
\vskip -12pt
\end{figure}

For all VAEs, we use the standard convolutional encoder and decoder with LeakyReLU \cite{leakyrelu_maas2013rectifier} activated strided convolutional and transposed convolutional blocks with BatchNorm \cite{ioffe2015_batchnorm}, respectively. This architectural setting allows various implementational advantages where convolutional accelerators and other ASICs can be used to speed up inference to achieve lower-latency and near-realtime performance. We resize the images to 64x64 and all VAEs use reconstruction as the primary task, where the loss function is given as:
\begin{equation}
    \text{L}(x;\phi, \theta) 
			= -\mathbb{E}_{z\sim q_{\phi}(z|x)}[\log p_{\theta}(x|z)] + \text{D}_{\text{KL}}(q_{\phi}(z|x)||p_{\theta}(z))  \\
\end{equation}
where $x$ is the input image, $z$ is the latent vector, $\phi$ and $\theta$ denote parameters of the encoder and decoder, respectively, and $D_{\text{KL}}$ is the Kullback-Leibler divergence. Note that for each VAE variant, the corresponding loss will also contain additional specialized loss terms. The reader is referred to the associated paper for each variant for further details.

We train all VAEs at a constant learning rate of 0.005 and no weight decay up to a maximum of 500 epochs with an early stopping criteria set on validation loss with a patience of 100 epochs, i.e, if there is no improvement for the past 100 epochs, the training ends. Once the training is complete, we check the reconstructed images manually as shown in \autoref{fig:vae_recons}. DIPVAE \cite{dipvae_kumar2017variational} produces reasonably good reconstructions among all other VAEs, and the reconstructions evidence that it captures coarse information necessary to understand the scene. We hypothesize that DIPVAE reconstructions are less blurry as disentanglement is encouraged by introducing a regularizer over the induced inferred prior. $\beta$-VAE \cite{betavae_DBLP:conf/iclr/HigginsMPBGBML17} also encourages disentanglement, but with DIPVAE there is no extra conflict introduced between disentanglement of the latents and the observed data likelihood. Further, we additionally trained a DIPVAE on 128x128 images, which yielded similar reconstruction results.

Once trained, we use the VAE encoder to infer latent vectors for input images to use as compact global descriptor embeddings. As such, the decoder module of the VAE contributes to loss during training and is not used during inference. Nonetheless, the decoder may still be used to reconstruct images from latent vectors which facilitates interpreting and visualizing the global descriptors. The encoder is confined to standard normal distribution and hence, this makes it easier to tweak latent vectors and to visualize their corresponding reconstructions. This makes it possible to understand the feature or variation encoded in required dimensions using the decoded image. Additionally, DIPVAE encourages disentangling of features in the latent space, which results in less overlap of variations across the dimensions, producing more meaningful and interpretable intermediate representations to use as global descriptors. Ultimately, such configurations producing intermediate representations instead of directly mapping pixels to actions, are known to achieve higher task performance, and generalize better to previously unseen environments \cite{intermediateRepresentations_zhou2019does}.

\subsection{Scene Classification}
We train a linear classifier on top of the frozen base VAE network on the train split of the dataset for 100 epochs. A two-layer input-output perceptron without any activation function is used as the linear classifier.

\label{Dataset}

\begin{wraptable}{r}{0pt}
\centering
\begin{tabular}{lrlrlr}
\hline
\multicolumn{2}{c}{\textbf{Rural}}           & \multicolumn{2}{c}{\textbf{Suburban}}        & \multicolumn{2}{c}{\textbf{Urban}} \\ \hline
Jarrahdale Perth  & \multicolumn{1}{r|}{33}  & Hawaii           & \multicolumn{1}{r|}{8}  & Indianapolis         & 30          \\
Missouri Ozarks   & \multicolumn{1}{r|}{22}  & Howth            & \multicolumn{1}{r|}{17} & Nashville            & 21          \\
Southern Illinois & \multicolumn{1}{r|}{43}  & Melbourne        & \multicolumn{1}{r|}{33} & Paris                & 24          \\
Stockport Buxton  & \multicolumn{1}{r|}{25}  & Stockport Buxton & \multicolumn{1}{r|}{17} & St Louis             & 52          \\
Utah              & \multicolumn{1}{r|}{75}  & Wimbledon        & \multicolumn{1}{r|}{21} & Toronto              & 33          \\ \hline
Rural             & \multicolumn{1}{r|}{198} & Suburban         & \multicolumn{1}{r|}{96} & Urban                & 160         \\ \hline
\multicolumn{3}{l}{\textbf{Total}}                              & \multicolumn{3}{r}{\textbf{454}}                             \\ \hline
\end{tabular}
\caption{Information about our Scene Categorization dataset}
\vskip 5pt
\label{table:dataset}
\end{wraptable}

We curate our own dataset\footnote{\label{footn:dataset}Dataset available to download from {\scriptsize \url{https://gist.github.com/saravanabalagi/1cda6ae06c4cf722fd2227e83eadc792}}} for training and testing the classifier, by manually selecting screenshots at different timestamps from driving videos on YouTube. This dataset includes 3 scene categories: Rural, Urban and Suburban, each category includes images captured in or around a diverse set of cities and regions as shown in \autoref{table:dataset}. Some examples are shown in \autoref{fig:scene_categorization} and as can be seen the dataset covers a variety of landscapes including desert and mountainous landscapes and exhibit mild to moderate illumination and seasonal changes such as fallen leaves, different time of the day, etc. The train split is made by randomly selecting two-thirds of the images from each route yielding 314 images and the linear classifier is trained.

\section{Experiments}
\label{Experiments}

To verify the suitability of the embeddings for scene categorization, we use the widely used evaluation procedure employed to test embeddings for classification tasks
\cite{Imagenet_5206848,MomentumContrast_9157636}.
Our proposed architecture already uses an intermediate global descriptor representation, which is then input to the linear classifier for scene categorization. Hence, we evaluate the resultant output without adding any additional layers.

\subsection{Evaluation}
The output of the linear classifier is tested on the test split of the dataset containing the remaining 140 images and the test accuracy is used as a proxy for representation quality of the embeddings used. Evaluation was done on an Intel i9-9900K (8 cores @3.60 GHz) and Nvidia RTX 2080 Ti and all images were resized to 128x128.
We compare the results with benchmark learned and handcrafted holistic image descriptors: (1) NetVLAD\footnote{MATLAB implementation provided by authors at {\scriptsize\url{https://github.com/Relja/netvlad}} is used} \cite{arandjelovic2016netvlad}, a weakly supervised CNN with generalized VLAD (Vector of Locally Aggregated Descriptors) layer. (2) PHOG\footnote{Code extracted from C++ implementation provided by authors at {\scriptsize\url{https://github.com/emiliofidalgo/htmap}} is used} \cite{phog_10.1145/1282280.1282340}, Pyramid Histogram of Gradients. 
For the evaluation we consider the following candidates:
\begin{itemize}
    \itemsep -4pt
    \item NetVLAD 4096 dimensions: Supervised, pretrained on Pittsburgh dataset\footnote{\label{footn:netvlad_downloads}Off-the-shelf VGG16+NetVLAD+whitening model provided at {\scriptsize \url{https://www.di.ens.fr/willow/research/netvlad/}}}
    \item NetVLAD 128 dimensions: Supervised, pretrained, same as above, cropped to 128 dimensions and L2-normalized from NetVLAD 4096 embedding
    \item PHOG 1260 dimensions: Handcrafted, 60 bins and 3 levels \cite{htmap_7938750}
    \item DIPVAE\footnote{Our own implementation in Python 3.8 and PyTorch 1.11 (CUDA 11.3) is used} 
    128 dimensions: Unsupervised, pretrained on 128x128 Oxford Robotcar dataset images
\end{itemize}

The experimental results are shown in \autoref{table:val_acc}. As expected, the supervised techniques scores higher than the unsupervised and handcraft techniques. NetVLAD 4096 tops the evaluation with 99.29\% accuracy, followed by the NetVLAD 128 with 94.29\% accuracy. The high accuracy is the result of (1) the technique's use of supervised learning, (2) the embedding length of 4096 allows capturing more information about the scene, and, (3) NetVLAD Cropped (128 dimensions) is computed from NetVLAD 4096 by cropping and normalizing.
DIPVAE (128 dimensions) achieves 82\% accuracy while only using 10\% embedding size as that of PHOG (1260 dimensions) and 3.1\% embeddings size as that of NetVLAD (4096 dimensions). 
We note that both NetVLAD and DIPVAE use GPU acceleration, while PHOG uses CPU optimizations and multi-threading. DIPVAE is computed more than twice as fast as PHOG and several orders of magnitude faster than NetVLAD.

\begin{wraptable}{r}{0pt}
\centering
\begin{tabular}{lrrrr}
\hline
\textbf{Route} & \textbf{Dublin} & \textbf{Vancouver} & \textbf{Wicklow} & \textbf{Redwood} \\ \hline
Total Images   & 14677           & 64672              & 60509            & 45253            \\ 
Test Images    & 11022           & 51018              & 47688            & 35483            \\ \hline
NetVLAD 4096   & 93.37           & 96.99              & 99.99            & 99.86            \\
NetVLAD 128    & 78.47           & 98.84              & 99.98            & 99.87            \\
PHOG 1260      & 91.25           & 98.50              & 88.46            & 86.60            \\
DIPVAE 128     & 95.70           & 83.72              & 99.44            & 95.60            \\ \hline
\end{tabular}
\caption{Accuracy (in \%) on the Scene Categorization Video Dataset}
\vskip 5pt
\label{tab:video_results}
\end{wraptable}

We further evaluate the linear classifiers on a second video based dataset\textsuperscript{\ref{footn:dataset}}. Here, we utilised frames from a variety of extended driving videos collected from YouTube as shown in \autoref{tab:video_results}. Each video was labelled as a single scene category, where collectively this resulted in a total of over 185K images. On each route, we remove the first 900 frames (30 seconds at 30 fps) to avoid encountering intro text, crossfades and other effects,  and use first 20\% of the frames ($\sim$40K) for training and the rest 80\% ($\sim$145K) for evaluation. We note that some portios of the  
sequences may exhibit ambiguous scene types and hence there will be a consequent noise in the results e.g. some images from urban sequences driven may resemble suburbs or rural regions. 
However given the length of each sequence, we estimate all Rural (Wicklow and Redwood) and City (Dublin and Vancouver) videos contain at least three quarters of images are unambiguously mapped to \textit{Rural} and \textit{Urban} labels, respectively. Therefore, we consider a model to be demonstrating good performance if it scores above 75\%.
\autoref{tab:video_results} shows the accuracy of the descriptors on videos of each region or city. As such, DIPVAE performs consistently well and shows similar performance to that of NetVLAD and PHOG while having much smaller embedding dimensionality and significantly faster compute time.

\begin{table}
\centering
\begin{tabular}{lcccrr}
\hline
\textbf{Descriptor} & \textbf{Type} & \textbf{Dimensions} $\downarrow$ & \textbf{Accuracy} (\%) $\uparrow$ & \textbf{Compute Time ($\mu$s)} $\downarrow$ \\ \hline
Random                                        & Trivial                                 & 4096                                          & 34.29                                       & 71.3 ± 0.0          \\
Random                                        & Trivial                                 & 128                                           & 28.57                                       & 2.7 ± 0.0           \\ \hline
NetVLAD                                       & Supervised                              & 4096                                          & \textbf{99.29}                              & 27560.0 ± 230.2     \\
NetVLAD Cropped                               & Supervised                              & \textbf{128}                                  & 94.29                                       & 27563.9 ± 230.4     \\ \hline
PHOG                                          & Hand-crafted                            & 1260                                          & 84.29                                       & 123.6 ± 3.9         \\
DIPVAE (Ours)                                 & Unsupervised                            & \textbf{128}                                  & 82.86                                       & \textbf{60.4 ± 3.0} \\ \hline
\end{tabular}
\caption{Test Accuracy reported on Scene Categorization Dataset for different descriptors mentioned in \Cref{Dataset}.
Random descriptor, constructed trivially by sampling numbers from normal distribution, is shown in top provide a baseline for trivial descriptors that do not capture any relevant information from the images. Compute time is the time taken to obtain the descriptor from a decoded image loaded in memory.}
\label{table:val_acc}
\end{table}


\section{Conclusion and Future Work}
\label{Future Work}

Our proposed solution to scene categorization uses an architecture made up of an unsupervised VAE-based embedding generator and a supervised light linear classifier head, and produces meaningful and interpretable intermediate representations as opposed to an end-to-end pixel-to-class approach with no explicit intermediate representations. 
The experimental results evidence that the DIPVAE latent vectors capture coarse information from the image, supporting their usage as global descriptors. These global descriptors exist in the multi-dimensional standard normal manifold, allowing easier comparison and interpretation compared to unbounded embedding hyperspaces. The proposed global descriptor is very efficient with a compact embedding length of 128, significantly faster to compute, and is robust to seasonal and illuminational changes, while capturing sufficient scene information required for scene categorization. Further, the VAE backbone's architecture made up of standard convolutional blocks allows more efficient, fast, low latency and near-realtime inference using hardware convolutional accelerators, substantiating their use in autonomous vehicles to quickly determine location context as precursor task. We note that there is a potential to further improve this performance using supervised and weakly supervised techniques. If and when labels are available, the VAE backbone can be set to learn with a small learning rate (e.g.
one-tenth relative to that of the head) in an end-to-end manner.
Additionally, further available information, such as GPS, together with recent predictions, could also be used to make more temporally consistent decisions about the scene category (e.g. avoiding categorising the environment as rural when driving along a tree-lined route in a city). Finally, we indicate that the proposed global descriptors, being intermediate representations, are useful for other tasks and actions that only require coarse features including scene information present in the image.

In our future work we intend to explore the potential of adding further categories such as motorways, tunnels, car-parks, etc. that are useful and provide more context for various autonomous driving tasks. Given the successful results, we further intend to integrate this approach in a hierarchical place recognition pipeline, where these compact global representations are used to aggregate images to facilitate faster image retrieval.


\bibliographystyle{apalike}
\bibliography{root}

\end{document}